\documentclass[lettersize,journal]{IEEEtran}
\usepackage{amsmath,amsfonts}
\usepackage{algorithmic}
\usepackage{array}
\usepackage[caption=false,font=normalsize,labelfont=sf,textfont=sf]{subfig}
\usepackage{textcomp}
\usepackage{stfloats}
\usepackage{threeparttable}
\usepackage{booktabs}
\usepackage{graphicx}
\usepackage{cleveref}
\usepackage{multirow}

\usepackage{url}
\usepackage{verbatim}
\usepackage{graphicx}
\usepackage[linesnumbered,ruled,vlined]{algorithm2e}

\def\BibTeX{{\rm B\kern-.05em{\sc i\kern-.025em b}\kern-.08em
    T\kern-.1667em\lower.7ex\hbox{E}\kern-.125emX}}
\usepackage{balance}
\begin{document}
\title{ConsistencyDet: A Few-step Denoising Framework for Object Detection Using the Consistency Model}
\author{Lifan~Jiang,~
        Zhihui~Wang,~
        Changmiao~Wang,~
        Ming~Li,~\IEEEmembership{Member,~IEEE,}
        Jiaxu~Leng
\thanks{Lifan Jiang, Zhihui Wang are with the College of Computer Science and Engineering, Shandong University of Science and Technology, Qingdao 266510, China (e-mail: lifanjiang@sdust.edu.cn;zhihuiwangjl@gmail.com;).}
\thanks{Changmiao Wang is with Shenzhen Research Institute of Big Data, Shenzhen, 518172, China (e-mail: cmwangalbert@gmail.com).}
\thanks{Ming Li is with Zhejiang Key Laboratory of Intelligent Education Technology
and Application, Zhejiang Normal University, Jinhua 321004, China (e-mail: mingli@zjnu.edu.cn).}
\thanks{Jiaxu Leng is with the Key Laboratory of Image Cognition, Chongqing University of Posts and Telecommunications, Chongqing 400065, China, and also with Jiangsu Key Laboratory of Image and Video Understanding for Social Safety, Nanjing University of Science and Technology, Nanjing 210094, China (e-mail: lengjx@cqupt.edu.cn).}
}

\markboth{xxxx xxxx,~Vol.~xx, No.~xx, xx~xxxx}{}%

\maketitle

\begin{abstract}
Object detection, a quintessential task in the realm of perceptual computing, can be tackled using a generative methodology. In the present study, we introduce a novel framework designed to articulate object detection as a denoising diffusion process, which operates on the perturbed bounding boxes of annotated entities. This framework, termed \textbf{ConsistencyDet}, leverages an innovative denoising concept known as the Consistency Model. The hallmark of this model is its self-consistency feature, which empowers the model to map distorted information from any time step back to its pristine state, thereby realizing a \textbf{``few-step denoising''} mechanism. Such an attribute markedly elevates the operational efficiency of the model, setting it apart from the conventional Diffusion Model. Throughout the training phase, ConsistencyDet initiates the diffusion sequence with noise-infused boxes derived from the ground-truth annotations and conditions the model to perform the denoising task. Subsequently, in the inference stage, the model employs a denoising sampling strategy that commences with bounding boxes randomly sampled from a normal distribution. Through iterative refinement, the model transforms an assortment of arbitrarily generated boxes into definitive detections. Comprehensive evaluations employing standard benchmarks, such as MS-COCO and LVIS, corroborate that ConsistencyDet surpasses other leading-edge detectors in performance metrics. Our code is available at \url{https://anonymous.4open.science/r/ConsistencyDet-37D5}.
\end{abstract}


\section{Introduction}
\label{sec:intro}

\IEEEPARstart{O}{bject} detection is a fundamental task in computer vision that predicts both positional data and categorical identities for objects in images \cite{10028728,10089211,10168277}. It supports various applications, including instance segmentation, pose estimation, and action recognition \cite{9693155,10041022,6942210,ZHANG2024110249,10098642}. Early methods relied on sliding windows and region proposals for defining candidate regions, using surrogate techniques for regression and classification \cite{girshick2015fast,ren2015faster}. While effective, these approaches were limited by manual candidate selection, affecting adaptability in complex contexts. The introduction of anchor boxes enhanced flexibility for diverse object scales and aspect ratios \cite{lin2017focal,redmon2016you}, although reliance on prior information restricted their applicability across datasets. Recently, the field has shifted towards learnable object queries or proposals \cite{carion2020end,sun2021sparse}, with significant advancements such as the Detection Transformer (DETR), which employs learnable queries to recast detection as an end-to-end learning problem \cite{carion2020end}. This approach eliminated the need for hand-crafted components and led to impressive results in various visual recognition tasks, especially in small object detection (SOD).

\begin{figure}[!t]
\centering
\includegraphics[width=\linewidth]{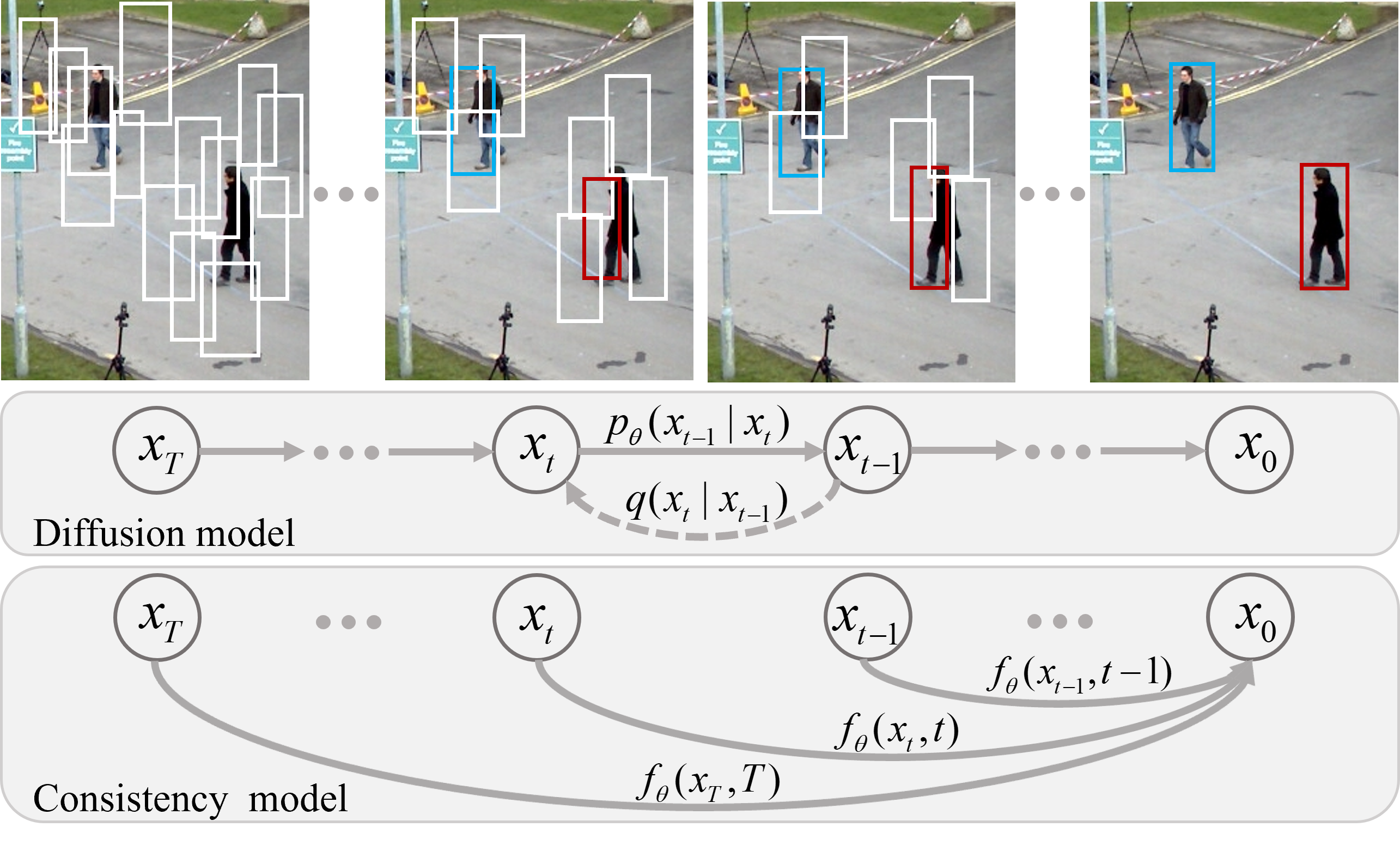}
\caption{Comparisons of denoising strategies of the Diffusion Model and Consistency Model for object detection. Object detection can be regarded as a denoising diffusion process from noisy boxes to object boxes. In the Diffusion Model, \(q(\cdot|\cdot)\) is the diffusion process and $ p_{\theta}(\cdot |\cdot) $ is the reverse process with a stepwise denoising operation. In the Consistency Model, $ f_{\theta}(\cdot,\cdot) $ represents a one-step denoising process.}
\label{fig:1}
\end{figure}

Diffusion Model \cite{dhariwal2021diffusion,10081412}, a score-based generative model, has shown remarkable effectiveness in image generation \cite{10105619}, segmentation \cite{brempong2022denoising}, and object detection \cite{chen2023diffusiondet}. Its iterative sampling mechanism systematically reduces noise from an initially random vector. In object detection, this process begins with stochastic bounding boxes, refining accuracy through iterations to delineate objects effectively. This ``noise-to-box'' methodology obviates the need for heuristic priors, simplifying object candidate selection. Based on Diffusion Model principles, DiffusionDet has outperformed existing detectors, even surpassing the Transformer model paradigm \cite{chen2023diffusiondet}.

\begin{figure}[!t]
\centering
\includegraphics[width=0.9\linewidth,height=170px]{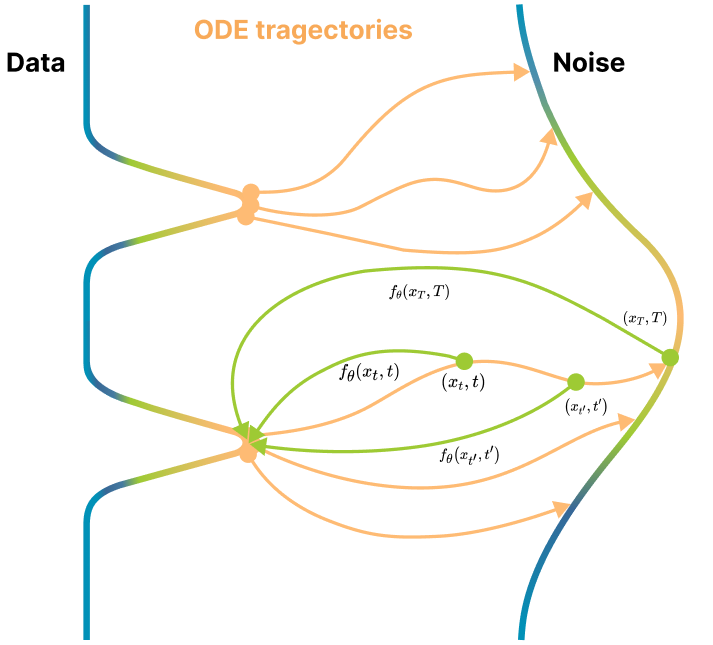}
\caption{Consistency Model undergoes training process to establish a mapping that brings points along any trajectory of the PF ODE back to the origin of that trajectory \cite{song2023consistency}.}
\label{fig:2}
\end{figure}

However, its incremental denoising process limits flexibility and computational efficiency, requiring further optimization for practical applications. We propose an innovative framework, \textbf{ConsistencyDet}, building on the foundational principles of DiffusionDet \cite{song2023consistency}. As shown in Fig. \ref{fig:1}, a comparative analysis reveals that the self-consistency property of the Consistency Model enables few-step denoising, significantly improving computational efficiency while reducing the required iteration count without sacrificing detection accuracy. In Fig. \ref{fig:2}, our approach leverages the ordinary differential equation (ODE) framework for probability flow (PF) from DiffusionDet \cite{song2020score}, facilitating a smooth transition from data distributions to noise distributions. ConsistencyDet uniquely maps points from arbitrary time steps back to their trajectory origin, enabling efficient sample generation by transforming random noise vectors into structured outputs with minimal network evaluations.

Furthermore, ConsistencyDet effectively integrates Gaussian noise \cite{ho2020denoising} into the coordinates and sizes of bounding boxes, creating noisy boxes for feature extraction from regions of interest (RoI) \cite{ren2015faster}. These RoI features, which are obtained from output maps produced by deep backbone networks like ResNet-50 \cite{he2016deep}, are then fed into a detection decoder to predict noise-free ground truth boxes. Consequently, ConsistencyDet can accurately infer the true bounding boxes from randomly generated ones, generating them through an inverse diffusion process with minimal computational cost and high efficiency.

The performance of ConsistencyDet has been rigorously evaluated on challenging datasets like MS-COCO and LVIS v1.0 \cite{lin2014microsoft,gupta2019lvis}, demonstrating commendable results across various backbones, including ResNet-50, ResNet-101, and Swin-Base \cite{he2016deep,liu2021swin}. Our contributions are as follows:
\begin{itemize}

\item Based on the self-consistency characteristic of the Consistency Model, in the design of the loss function, we accumulate the loss values at time steps $t$ and $t-1$ to compute the overall loss. This ensures that the mapping results of adjacent points along the temporal axis to the origin remain stable, thereby enhancing the model's ``few-step denoising" capability.

\item Our work introduces the principles of the Consistency Model to the task of object detection for the first time, proposing a novel detection method and establishing a well-designed pipeline for it, termed ConsistencyDet. Our work offers a new perspective and research direction for the field of object detection.

\item Based on the unique characteristic of self-consistency, ConsistencyDet enables the model to efficiently map distorted information from any temporal stage back to its original state, thereby achieving a highly effective ``few-step denoising" mechanism. Compared to other object detection methods based on the diffusion model, this distinctive feature significantly enhances the model's overall inference speed and computational efficiency.
\end{itemize}

\section{Related works and nomenclature}
\label{Related}
\subsection{Object detection}
\label{YOLOX}

Object detection has undergone significant evolution across various phases. In the initial stages, researchers employed traditional image processing techniques, endeavoring to detect objects by employing methods such as edge detection and feature extraction. These methods, however, encountered considerable challenges when dealing with complex scenes and variations in illumination. With the advent of deep learning, seminal frameworks such as R-CNN and Fast R-CNN \cite{girshick2015fast} have significantly enhanced detection capabilities in terms of speed and accuracy. This was achieved through the integration of innovative concepts like Region Proposal Networks and RoI Pooling. Subsequent developments led to the emergence of single-stage detectors, including YOLO and SSD, which offered considerable improvements in real-time performance, facilitating rapid end-to-end object detection. More recently, the introduction of attention-based Transformer methods, exemplified by DETR \cite{carion2020end}, has propelled object detection to unprecedented levels of performance, enabling more precise object localization by leveraging global visual context modeling. Although these detection methods have demonstrated noteworthy effectiveness, there remains considerable potential for further advancement. In the present work, we introduce an innovative detection approach that iteratively refines the position and size of bounding boxes. This iterative refinement process employs a series of noisy iterations, culminating in the bounding boxes precisely encompassing the target object.

\subsection{Diffusion Model for perception tasks}
\label{reid}
The Diffusion Model \cite{sohl2015deep,ho2020denoising,song2020denoising,song2020score} is a powerful and highly effective generative model that reconstructs data from random noise through a denoising process and has shown impressive success in various fields like computer vision \cite{10081412} and natural language processing \cite{austin2021structured}. In object detection, it has been adapted into DiffusionDet \cite{chen2023diffusiondet}, which reinterprets detection as a set prediction task by assigning object candidates to ground truth (GT) boxes. Building upon DiffusionDet, our work aims to further enhance detection efficiency by introducing a novel single-step method that preserves the benefits of iterative sampling. This approach optimizes the balance between detection accuracy and computational speed, providing a streamlined yet highly effective and computationally efficient solution for complex object detection tasks.

\begin{table}[htbp]
\setlength{\tabcolsep}{0.1pt}
\centering\scriptsize
\scalebox{1.02}{
\begin{threeparttable}
 \caption{\label{tab:test1_1}Nomenclature with related notations.}
 \begin{tabular}{p{2.0cm} p{6.5cm}}\toprule
\textbf{Notation\ }  & \textbf{Definition} \\
\cmidrule(r){1-2}
$T$ & Number of total time steps\\
$t$ & Current time step\\
$t_r$ & A random time step in the range $[0, T]$\\
$\Delta t$ & Time step interval for sampling\\
$\textit{c}_x^\textit{i}/\textit{c}_y^\textit{i}$ & $x/y$-axis coordinate of the $i$-th box's center point\\
$\textit{w}^\textit{i}/ \textit{h}^\textit{i}$  & Width \!/\! Height of the $i$-th box\\
$\textit{b}^\textit{i}$    & $(\textit{c}_x^\textit{i}, \textit{c}_y^\textit{i}, \textit{w}^\textit{i}, \textit{h}^\textit{i})$ of the $i$-th box\\
$\alpha_t/\sigma_t$         & Parameter in Denoiser at the $t$-th time step\\
$\theta$   & Model parameter \\
$F_{\theta}(\cdot,\cdot)$ & A designed free-form deep neural network\\
$\mathcal{N}(\cdot,\cdot)$  & Normal distribution\\
$f_{\theta}(\cdot,\cdot)$   & Final answer for Consistency Model\\
$c_{skip/out/in}(\cdot)$    & Calculation factor for $f_{\theta}$\\
$d(\cdot, \cdot)$  & Distance function\\
$\eta$ & Learning rate\\
$\mu$ & EMA decay rate\\
$\Phi(\cdot, \cdot; \phi)$ & ODE solver\\
$\lambda(\cdot)$  & A positive weighting function\\
$\mathcal{L}$ & Total loss function in training phase\\
$\mathcal{L}_{cls/L1/giou}$ & Focal \!/\! L1 \!/\! GIoU loss item\\
$\lambda_{cls/L1/giou}$   & Weight for Focal \!/\! L1 \!/\! GIoU loss item\\
$\sigma_{max/min}$      & Maximum \!/\! Minimum threshold of noise parameter\\
$\sigma_{data}$   & Noise parameter between $\sigma_{min}$ and $\sigma_{max}$\\
$\epsilon$           & Randomly generated Gaussian noise\\
$\rho$    & Scale factor of generated noise\\
$B_t$ & Random noise at the $t$-th time step in sampling\\
$r(\cdot)$ & Generate random noise with given dimensions\\
$E(\cdot)$ & Image feature extraction with backbone network\\
$P_c(\cdot,\cdot)$ & Prediction of Consistency Model in each time step\\
$nms(\cdot,\cdot)$ & Non-max suppression (NMS) operation\\
$N_{th}$ & Threshold of NMS operation\\
$B_{th}$ & Threshold of Box-renewal operation\\
$dcm(\cdot,\cdot)$ & Decoder of ConsistencyDet with head network\\
$ddm(\cdot,\cdot)$ & Denoiser of DiffusionDet with head network\\
$conc(\cdot,\cdot)\ $\ & Concatenate function\\
$n_{ss}$ & Number of sampling steps\\
$n_{tr}$ & Number of total proposed boxes in training phase\\
$n_p$ & Number of total proposed boxes in inference\\
$n_r$ & Number of current proposed boxes\\
$x_s$ & Padded box information at time axis origin\\
$x_t$ & Noised box information at the $t$-th time step\\
$x_b$ & Predicted box information in each time step\\
$x_0$ & Predicted box information at time axis origin\\
$x_{box/cls}$ & Predicted object's box coordinate \!/\! category\\
AP    & Average Precision\\
AP$_{50/75}$ & Average Precision at 50\% \!/\! 75\% IoU\\
AP$_{s/m/l}$ & Average Precision for small \!/\! median \!/\! large objects\\
AP$_{r/c/f}$  & Average\! Precision\! for\! rare \!/\! common \!/\! frequent\! categories\\

\bottomrule
 \end{tabular}\label{tb:Nomenclature}
  \end{threeparttable}
}
\end{table}

\subsection{Consistency Model}
\label{cons_m}

The Diffusion Model is predicated on an iterative generation process, which tends to result in sluggish execution efficiency, thus curtailing its applicability in real-time scenarios. To address this limitation, OpenAI has unveiled the Consistency Model, an innovative category of generative models capable of rapidly producing high-quality samples without necessitating adversarial training regimes. The Consistency Model facilitates swift one-step generation, yet it retains the option for multi-step sampling as a means to navigate the trade-off between computational efficiency and the caliber of generated samples. Furthermore, it introduces the capability for zero-shot data manipulation \cite{2014Attribute}, encompassing tasks such as image restoration, colorization, and super-resolution, obviating the need for task-specific training. The Consistency Model can be cultivated through a distillation pre-training regimen derived from existing Diffusion Models or alternatively as a standalone generative model. This work formally acknowledges this capability and, for the inaugural time, incorporates the Consistency Model within the domain of object detection, hereby designated as ConsistencyDet.

\subsection{Nomenclature}
\label{Nome}

For the sake of clarity in the ensuing discussion, we provide a summary of the symbols and their corresponding descriptions as utilized in this study. This is encapsulated in Table \ref{tb:Nomenclature}, which meticulously outlines the nomenclature employed. The symbols encompass a variety of elements including training samples, components of the loss function, strategies for training, and metrics for evaluation, among others.

\section{Proposed detection method}
\label{Proposed}
In this section, we first introduce the core mathematical principles and underlying theoretical foundations of our method in~\cref{Preliminaries}. Then, we provide a comprehensive and detailed description of the proposed pipeline in~\cref{Loss}, which effectively integrates the principles of the Consistency Model into the object detection task. Next, we will delve deeper into a thorough discussion of the specific training strategies and methodologies involved in ConsistencyDet, as shown in~\cref{Foreground}. Finally, we thoroughly examine the complete inference process undertaken by ConsistencyDet in~\cref{Location}, including all relevant steps and considerations.

\subsection{Preliminaries}
\label{Preliminaries}
\noindent
\textbf{Object Information.}
In the domain of object detection, datasets consist of input-target pairs denoted as ($\mathit{x}$, $\mathit{b}$, $\mathit{c}$), where $\mathit{x}$ represents an input image, $\mathit{b}$ denotes the set of bounding boxes for objects within the image, and $\mathit{c}$ signifies the corresponding set of category labels for those objects. More precisely, the $i$-th bounding box in the set $\mathit{b}$ can be quantified as $\mathit{b}^i = (\mathit{c}_x^i, \mathit{c}_y^i, \mathit{w}^i, \mathit{h}^i)$, where the coordinates $(\mathit{c}_x^i, \mathit{c}_y^i)$ specify the center of the bounding box along the x-axis and y-axis, respectively, while the terms $(\mathit{w}^i, \mathit{h}^i)$ define the width and height of the bounding box.

\vspace{0.3em}
\noindent
\textbf{Diffusion Model.}
Diffusion Models can be principally categorized into two types: Denoising Diffusion Probabilistic Models (DDPM) \cite{ho2020denoising} and Denoising Diffusion Implicit Models (DDIM) \cite{song2020denoising}. DDIM is an optimized variant of DDPM, where the generation process is modified. For DDIM, the procedure commences by predicting \(x_0\) from \(x_t\) and subsequently inferring \(x_{t-1}\) from \(x_0\). Here, \(x_0\) acts as an anchor point, enabling the generation process to navigate through arbitrary time steps, thus circumventing the temporal constraints intrinsic to DDPM. While the diffusion process and the training methodology of DDIM parallels that of DDPM, the sampling process in DDIM ceases to be a Markov chain because \(x_{t-1}\) is contingent not only on \(x_t\) but also on \(x_0\).

At any given moment, \(x_0\) can be derived from \(x_t\) and \(\epsilon\) iteratively, using the following relationship:
\begin{equation}\label{func1}
f_{\theta}^{(t)}(x_t) = \frac{x_t - \sqrt{1-\alpha_t} \cdot \epsilon_{\theta}^{(t)}(x_t)}{\sqrt{\alpha_t}}.
\end{equation}

Subsequently, \(x_{t-1}\) is predicted based on the neural network's output as per the equation:
\begin{equation}\label{func2}
p_{\theta}^{(t)}(x_{t-1}|x_t) =
\begin{cases}
\mathcal{N}(f_{\theta}^{(1)}(x_1), \sigma_1^2\mathbf{I}), & \text{if } t=1, \\
q_{\sigma}(x_{t-1}|x_t, f_{\theta}^{(t)}(x_t)), & \text{otherwise}.
\end{cases}
\end{equation}

When \(x_0\) and \(x_{t-1}\) (for \(t > 1\)) are given, the forward process is rendered deterministic. The generation of latent variable samples employs a consistent strategy, designated as the DDIM. The fundamental principle of DDIM is encapsulated in the equation:
\begin{equation}\label{func3}
\begin{split}
x_{t-1} = & \sqrt{\alpha_{t-1}} \left( \frac{x_t - \sqrt{1-\alpha_t} \cdot \epsilon_{\theta}^{(t)}(x_t)}{\sqrt{\alpha_t}} \right) \\
& + \sqrt{1 - \alpha_{t-1} - \sigma_t^2} \cdot \epsilon_{\theta}(x_t, t) + \sigma_t\epsilon_t.
\end{split}
\end{equation}

In this formulation, DDIM presents a refined approach to generating samples, affording a significant enhancement in efficiency over the traditional DDPM framework.

\vspace{0.3em}
\noindent
\textbf{Consistency Model.}
Within the framework of the Consistency Model which utilizes deep neural networks, two cost-effective methodologies are investigated for enforcing boundary conditions. Let \(F_{\theta}(x, t)\) represent a free-form deep neural network whose dimensionality is the same as \(x\). The first method directly parameterizes the Consistency Model as:

\begin{equation}\label{func4}
f_{\theta}(x, t) =
\begin{cases}
  x, & \text{if } t = \tau, \\
  F_{\theta}(x, t), & \text{if } t \in [\tau, T),
\end{cases}
\end{equation}
where \(\tau\) is an integer in the range \([0, T-1]\). The second method parameterizes the Consistency Model by incorporating skip connections and is formalized as follows:

\begin{equation}\label{func5}
f_{\theta}(x, t) = c_{skip}(t)x + c_{out}(t)F_{\theta}(x, t),
\end{equation}
where \(c_{skip}(t)\) and \(c_{out}(t)\) are differentiable functions \cite{song2023consistency}, satisfying \(c_{skip}(\tau) = 1\) and \(c_{out}(\tau) = 0\). By employing this construction, the Consistency Model becomes differentiable at \(t = \tau\), provided that \(F_{\theta}(x, t)\), \(c_{skip}(t)\), and \(c_{out}(t)\) are all differentiable. This differentiability is crucial for the training of continuous-time Consistency Models.




\begin{figure*}[!t]
\centering
\includegraphics[width=7.0in]{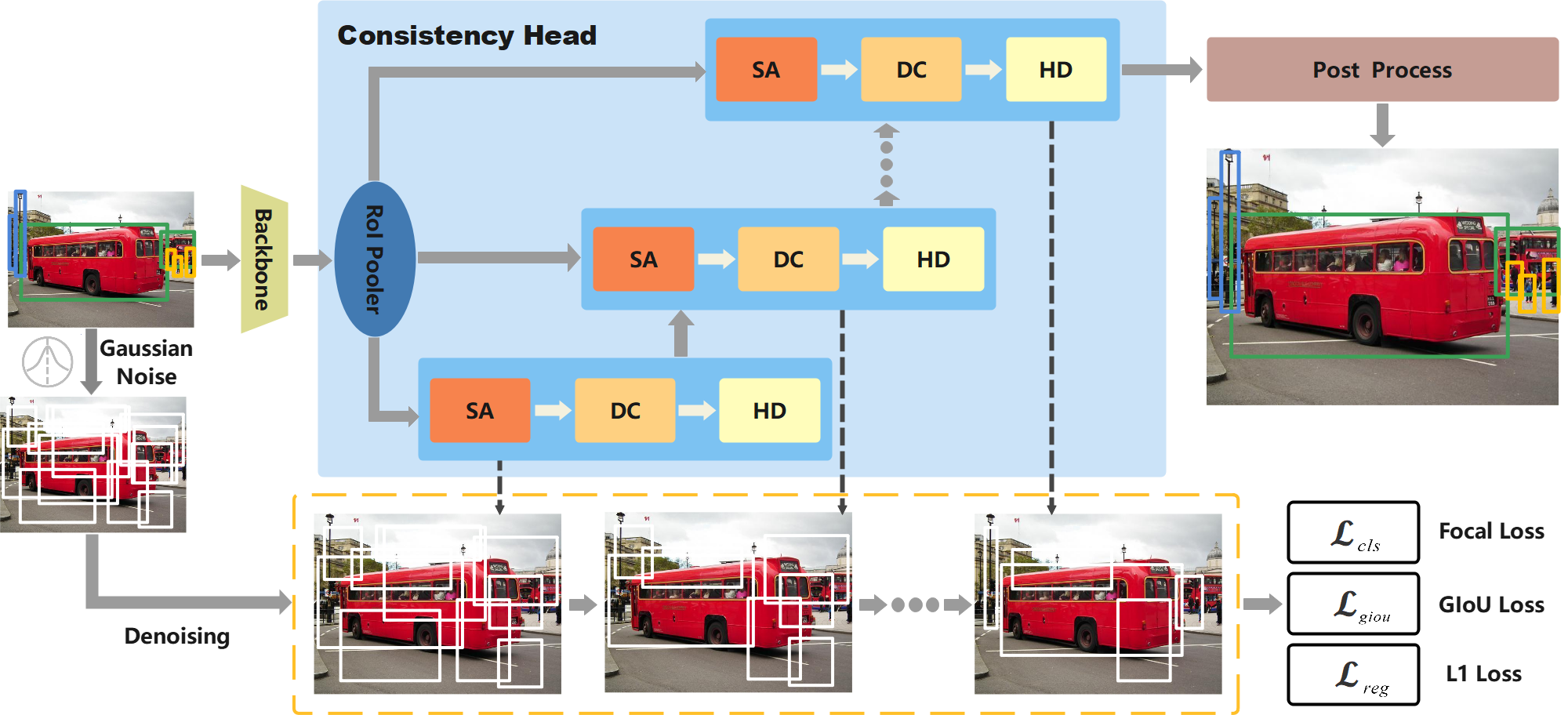}
\caption{Training procedures of the proposed ConsistencyDet. After extracting features through the backbone, random Gaussian noise is added to GT boxes following the Consistency Model's noise addition strategy. These noised boxes with corresponding features processed by RoI pooler are then input to ConsistencyHead for iterative noise removal, with several basic modules, ultimately yielding the final detection results. Each the basic module contains a self-attention mechanism (SA), dynamic convolutional layers (DC) and the head (HD) of classification and regression.}
\label{fig:feature_extraction}
\end{figure*}

\begin{figure}[!t]
\centering
\includegraphics[width=3.5in]{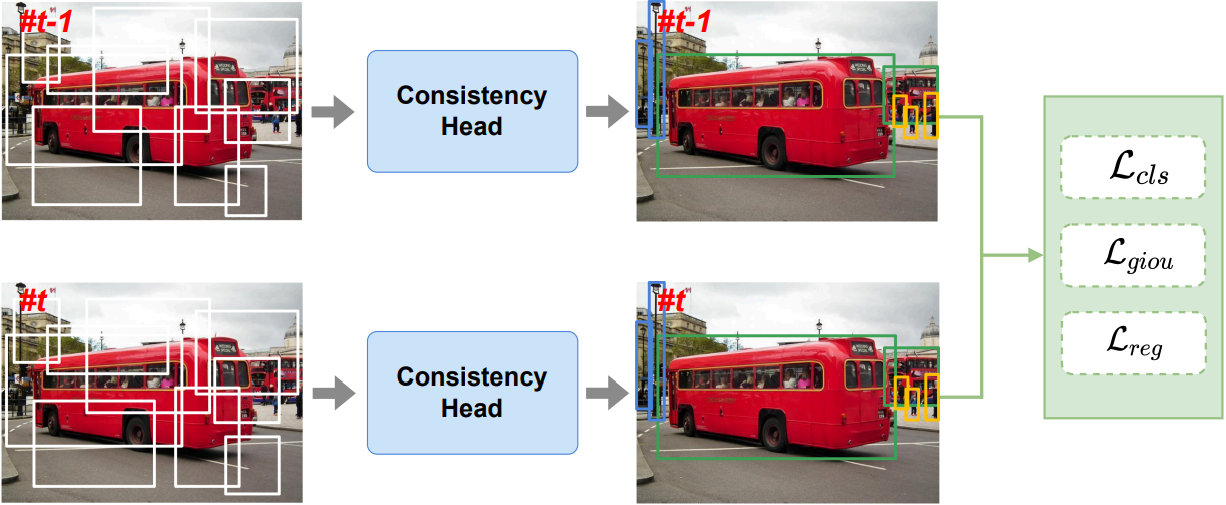}
\caption{To ensure the self-consistency property of the Consistency Model, it requires feeding the noised boxes corresponding to $(t-1)$-th and $t$-th time steps into the model simultaneously. They jointly predict the final results, then compare them with GT to estimate the training loss.}
\label{fig:feature_comparison}
\end{figure}

\vspace{0.3em}
\noindent
\textbf{Training process of Consistency Model.}
Within the context of the Consistency Model, this study introduces two different training strategies that capitalize on the self-consistency property. The proposed strategies encompass distillation training, which transfers knowledge from a Diffusion Model to the Consistency Model, and an autonomous training regime where the Consistency Model is trained independently. The first method employs numerical ODE solvers in conjunction with a pre-trained diffusion-based detector, referred to as DiffusionDet, to generate pairs of proximate points along a PF ODE trajectory. By minimizing the divergence between the model predictions for these point pairs, knowledge from the Diffusion Model is effectively transferred to the Consistency Model. As a result, this allows for the generation of high-quality samples through a single evaluation of the network \cite{2015Distilling,2019Distilling}. The comprehensive training protocol for the Consistency Model through distillation is encapsulated in Algorithm \ref{alg:1}. Conversely, the second approach dispenses with the need for a pre-trained Diffusion Model, thereby enabling the Consistency Model to be trained in a standalone manner. This technique delineates the Consistency Model as an independent entity within the broader spectrum of its applications. The detailed training process for the independently trained Consistency Model is encapsulated in Algorithm \ref{alg:2}.







\begin{algorithm}
\caption{Consistency Distillation (CD)}
\label{alg:1}
\KwIn{Dataset $\mathcal{D}$, initial model parameter $\theta$, learning rate $\eta$, ODE solver $\Phi(\cdot, \cdot; \phi)$, distance function $d(\cdot, \cdot)$, temperature function $\lambda(\cdot)$, and $\mu$}
\KwOut{Final trained model parameter $\theta$}
\BlankLine

$\theta^{-} \leftarrow \theta$

\While{not convergence}{
    Sample $x \in \mathcal{D}$ and $n \in U[1, N - 1]$\;
    Sample $x_{\sigma_{t+1}} \in \mathcal{N}(x, \sigma_{t+1}^2I)$\;
    $\bar{x}_{\sigma_t}^{\phi} \leftarrow x_{\sigma_{t+1}} + (\sigma_t - \sigma_{t+1}) \Phi(x_{\sigma_{t+1}}, \sigma_{t+1}; \phi)$\;
    $\mathcal{L}(\theta,\theta^-;\phi) \leftarrow \lambda({\sigma_t}) d(f_{\theta}(x_{\sigma_{t+1}}, \sigma_{t+1}), f_{\theta^-}(\bar{x}_{\sigma_t}^{\phi}, \sigma_t))$\;
    $\theta \leftarrow \theta - \eta \nabla_{\theta} \mathcal{L}(\theta, \theta^-; \phi)$\;
    $\theta^{-} \leftarrow \text{stopgrad}(\mu \theta^{-} + (1 - \mu) \theta)$\;
    \Indm
}
\Return $\theta$
\end{algorithm}

\begin{algorithm}
\caption{Consistency Training (CT)}
\label{alg:2}
\KwIn{Dataset $\mathcal{D}$, initial model parameter $\theta$, learning rate $\eta$, step schedule $T(\cdot)$, EMA decay rate schedule $\mu(\cdot)$, $d(\cdot,\cdot)$, and $\lambda(\cdot)$}

\KwOut{Updated model parameter $\theta$}
\BlankLine

$\theta^- \gets \theta$ and $k \gets 0$

\While{not convergence}{
    Sample $\mathbf{x} \in \mathcal{D}$, and $t \in \mathcal{U}\left[1, T(k) - 1\right]$;

    Sample $\epsilon \in \mathcal{N}(0, I)$;

    $ \mathcal{L}(\theta, \theta^-) \gets \lambda(\sigma_t) d( f_{\theta}\left(\mathbf{x} + \epsilon\sigma_{t+1}, \sigma_{t+1}\right), f_{\theta^-}\left(\mathbf{x} + \epsilon\sigma_t, \sigma_t\right) ) $;

    $\theta \gets \theta - \eta\nabla_{\theta}\mathcal{L}(\theta, \theta^-)$;

    $\theta^- \gets \text{stopgrad}(\mu(k)\theta^- + (1-\mu(k))\theta)$;

    $k \gets k + 1$;
}
\Return $\theta$
\end{algorithm}

\subsection{Pipeline}
\label{Loss}

ConsistencyDet is composed of an image encoder and a detection decoder, with the workflow depicted in Fig. \ref{fig:feature_extraction}. The input to the image encoder is the original image along with its GT boxes. These GT boxes are then subjected to a noise injection procedure as part of the Consistency Model, where Gaussian noise is randomly introduced. The perturbed boxes are processed by the RoI Pooler, which extracts features from the regions of candidate bounding boxes based on the overall image features extracted by the backbone network. These extracted features subsequently pass through a self-attention mechanism (SA) and dynamic convolutional layers (DC), which serve to refine the target features further. Temporal information is then utilized to adjust the target features. The refined features are subsequently input into the heads (HD) for classification and regression to determine the category probabilities of objects and predict bounding boxes. This step involves a gradual refinement of the positions and scales of the noisy boxes to converge on the predicted outcomes. In the post-processing module, the predicted results undergo filtering via Non-Maximum Suppression (NMS) and are rescaled to match the original image resolution.

\vspace{0.3em}
\noindent
\textbf{Image encoder.}
In the design of ConsistencyDet, three different backbone architectures are chosen to function as image encoders: ResNet-50 \cite{he2016deep}, ResNet-101 \cite{he2016deep}, and Swin Transformer-base \cite{liu2021swin}. To facilitate multi-scale representation, which is critical for capturing details across different object sizes, the Feature Pyramid Network \cite{lin2017feature} is integrated with each of these backbone networks.

\vspace{0.3em}
\noindent
\textbf{Detection decoder.}
The detection decoder proposed in this study is composed of several cascaded stages that feature specially designed basic modules. Typically, the number of these modules is set to six during the training phase, but this can be adjusted during inference. This architecture inherits its framework from DiffusionDet \cite{chen2023diffusiondet}, which is characterized by initiating the detection process with randomly initialized bounding boxes and requiring only those proposal boxes as input. Notably, it employs a detector head that is shared across iterative sampling steps and is orchestrated by time step embeddings \cite{amit2021segdiff}. This iterative mechanism distinguishes the detection decoder from those used in other methods, which generally operate in a single-pass fashion, thereby defining both the architecture and the methodology unique to ConsistencyDet.

The differences between the decoder proposed here and the one used in DiffusionDet can be summarized as follows:

\begin{itemize}
    \item The noised boxes fed into the detection decoder are not only subject to operations for the addition of Gaussian noise but are also scaled by a coefficient, \(c_{in}(t)/2\), to constrain the position of the noised boxes. In contrast, DiffusionDet applies a direct clamping operation on the Gaussian-noise-augmented input boxes.

    \item ConsistencyDet mandates that the noised boxes at each consecutive pair of time steps \((t\!\!-\!\!1, t)\) be processed by two different detection decoders to generate respective predictions. These predictions are then evaluated against the GT boxes to ensure the model's consistency. Notably, the noised boxes at the \((t\!\!-\!\!1)\)-th time step are not randomly generated; instead, they are derived through specific computational processes. In contrast, the Diffusion Model solely relies on randomly generated noised boxes at the current \(t\)-th time step.
\end{itemize}

\subsection{Training}
\label{Foreground}


During the training phase, the diffusion process is initialized by creating perturbed bounding boxes derived from the GT boxes. The objective of the model is then to reverse this perturbation, effectively mapping the noisy bounding boxes back to their original GT counterparts. Two different training methodologies are employed for ConsistencyDet. The first leverages extant weights from DiffusionDet, incorporating optional knowledge distillation as an enhancement. The second approach is a standalone method that relies on self-training, independent of DiffusionDet. The detailed procedures for each algorithm are delineated in Algorithms \ref{alg:3} and \ref{alg:4}. Notably in Algorithm \ref{alg:4}, should the knowledge distillation strategy be adopted, the denoiser references the detection outputs from the Diffusion Model. In the absence of knowledge distillation, the variable $x_s$ is directly utilized within the denoiser.

\begin{algorithm}[]
    \SetAlgoLined
    \DontPrintSemicolon
    \caption{Training loss of ConsistencyDet}
    \label{alg:3}

    \KwIn{Images $X$ with corresponding GT boxes, total time step $T$, a random time step $t_r$}
    \KwOut{Loss $\mathcal{L}_{t_r,t_{r-1}}$ per iteration}
    \BlankLine
    \For{\text{each iteration}}{
        Sample $X_{batch} \in X$;

        Extract features $E(X_{batch})$;

        Pad $X_{batch}$ with GT boxes and features as $x_s$;

        Generate a random time step $t_r$;

        \tcp{Calculate noise parameters}

        Calculate $(\sigma_{t_{r-1}}, \sigma_{t_r})$ by Eqn. (\ref{func6});

        Add noise to $x_s$ by Eqn. (\ref{func7}) as $x_{t_r}$;

        \tcp{Refer to Algorithm \ref{alg:4}}

        Predict $x_{t_{r-1}}$ with $(x_{t_r}, \sigma_{t_r}, \sigma_{t_{r-1}}, x_s)$;

        $d_{t_{r-1}} \gets dcm(x_{t_{r-1}}, \sigma_{t_{r-1}})$;

        $d_{t_r} \gets dcm(x_{t_r}, \sigma_{t_r})$;

        $\mathcal{L}_{t_r,t_r-1} \gets \mathcal{L}(d_{t_{r-1}},G)+\mathcal{L}(d_{t_r},G)$;

        \Return Loss $\mathcal{L}_{t_r,t_{r-1}}$\;
    }
\end{algorithm}

\vspace{0.3em}
\noindent
\textbf{GT boxes padding.}
In open-source benchmarks for object detection, such as those referenced in \cite{everingham2010pascal,gupta2019lvis,shao2018crowdhuman}, there is typically a variance in the number of annotated instances across different images. To address this inconsistency, we implement a padding strategy that introduces auxiliary boxes around the GT boxes. This ensures that the total count of boxes attains a predetermined number, $n_{tr}$, during the training phase. These padded instances are denoted as $x_s$, signifying the original padded samples. For the $i$-th GT box, represented by $\textit{b}^\textit{i}$, we apply Gaussian noise to its four parameters $(\textit{c}_x^\textit{i}, \textit{c}_y^\textit{i}, \textit{w}^\textit{i}, \textit{h}^\textit{i})$ at a randomly selected time step $t$.


\begin{algorithm}[]
    \SetAlgoLined
    \DontPrintSemicolon
    \caption{Denoiser with optional distillation}
    \label{alg:4}

    \KwIn{Noised box $x_t$, noise parameters $\sigma_t$ and $\sigma_{t-1}$, the origin padded $x_s$}
    \KwOut{Predicted box information $x_{t-1}$ at $(t\!-\!1)$-th time step}

    \BlankLine
    \eIf{no distillation}{
    $x_0 \gets x_s$;
    }
   {
    $x_0 \gets ddm(x_t, \sigma_t)$;
    }
    $\nabla_\sigma x \gets (x_t - x_0) / \sigma_t$;

    $x_{t-1} \gets x_t + \nabla_\sigma x \cdot (\sigma_{t-1} - \sigma_t)$;

    \Return $x_{t-1}$\;

\end{algorithm}

\vspace{0.3em}
\noindent
\textbf{Box corruption.} The range of the noised box at the $t$-th time step is constrained. Initially, the scale factor of the noise is determined as follows:

\begin{equation}\label{func6}
\sigma_{t} = \left(\sigma_{max}^{1/\rho} + \frac{t}{T - 1} \cdot \left(\sigma_{min}^{1/\rho} - \sigma_{max}^{1/\rho}\right)\right)^\rho.
\end{equation}

Subsequently, noise is introduced to the original padded sample $x_s$:

\begin{equation}\label{func7}
x_t = x_s + \epsilon \cdot \sigma_{t},
\end{equation}
where $\epsilon$ denotes randomly generated Gaussian noise.
Finally, the range of the noised box is restricted by:

\begin{equation}\label{fun9}
x_t \gets \frac{c_{in}(t)}{2} \cdot x_t,
\end{equation}
where $c_{in}(\cdot)$ represents the scale factor of the noised box and is defined as:

\begin{equation}\label{fun8}
c_{in}(t) = \frac{1}{\sqrt{{\sigma_{t}}^2 + \sigma_{data}^2}}.
\end{equation}

This formulation ensures that the noise scale factor is properly adjusted across the time steps, and that the noised box remains within the prescribed bounds.


\vspace{0.3em}
\noindent
\textbf{Loss function.} The loss functions employed to evaluate the predicted bounding boxes adhere to the framework established by DiffusionDet \cite{chen2023diffusiondet}, which incorporates both the $\mathcal{L}_{L1}$ loss and the $\mathcal{L}_{giou}$ loss. The former represents the standard \(L1\) loss, while the latter denotes the Generalized Intersection over Union (GIoU) loss as detailed by \cite{rezatofighi2019generalized}. Additionally, the classification of each predicted bounding box is assessed using the focal loss, denoted as $\mathcal{L}_{cls}$. To balance the relative influence of each loss component, a positive real-valued weight \(\lambda_{cls/L1/giou} \in \mathbb{R^+}\) is allocated to each term. Consequently, the aggregate loss function is formally expressed as:
\begin{equation}\label{fun10}
\mathcal{L} = \lambda_{cls} \cdot \mathcal{L}_{cls} + \lambda_{L1} \cdot \mathcal{L}_{L1} + \lambda_{giou} \cdot \mathcal{L}_{giou}.
\end{equation}

Leveraging the intrinsic attribute of self-consistency within the Consistency Model, the perturbed bounding boxes associated with a sample $x_s$ at consecutive time steps ($t\!-\!1$ and $t$) are subjected to a joint denoising process. The corresponding loss values are cumulatively computed to yield the final comprehensive loss:

\begin{equation}\label{fun10}
\begin{aligned}
\mathcal{L} = & \lambda_{cls} \cdot (\mathcal{L}_{{cls}_{t-1}} + \mathcal{L}_{{cls}_t}) \\
& + \lambda_{L1} \cdot (\mathcal{L}_{L1_{t-1}} + \mathcal{L}_{L1_t}) \\
& + \lambda_{giou} \cdot (\mathcal{L}_{{giou}_{t-1}} + \mathcal{L}_{{giou}_t}).
\end{aligned}
\end{equation}

\subsection{Inference}
\label{Location}
The inference mechanism implemented in ConsistencyDet bears resemblance to that of DiffusionDet, employing a denoising sampling strategy that progresses from initial bounding boxes, akin to the noised samples utilized during the training phase, to the final object detections. In the absence of ground truth annotations, these initial bounding boxes are stochastically generated, adhering to a Gaussian distribution. The model subsequently refines these predictions through an iterative process. Ultimately, the final detections are realized, complete with associated bounding boxes and category classifications. Upon executing all iterative sampling steps, the predictions undergo an enhancement procedure via a post-processing module, culminating in the final outcomes. The procedural specifics are delineated in Algorithm \ref{alg:5}.

\begin{algorithm}[]
    \SetAlgoLined
    \DontPrintSemicolon
    \caption{Inference of ConsistencyDet}
    \label{alg:5}
    \KwIn{Images $X$, total time step $T$, the number of sampling steps $n_{ss}$}
    \KwOut{Final predictions $nms(x_{box}, x_{cls})$}

    \BlankLine

    $\Delta t = T / n_{ss}$;

    \text{Generate random noise} $B_0$;

    Extract features $E(X)$;

    \For{$t = 0$ \KwTo $T-1$ \KwSty{step} $\Delta t$}{
    Calculate ${\sigma}_t$ by Eqn. (\ref{func6});

    $x_0, x_b, x_{cls}, x_{box}\gets P_c(E(X), B_t)$;

    Perform Box-renewal operation for $x_b$ and $x_0$;

    $\nabla_\sigma x \gets (x_b - x_0) / {\sigma}_t$;


    $B_t \gets x_b + \nabla_\sigma x (\sigma_{t + \Delta t} - \sigma_t)$;

    \tcp{Supplement new proposals}

    $B_t \gets conc(B_t, r([1, n_p - n_r, 4]) \cdot \sigma_{t + \Delta t})$;
    }

    \Return $nms(x_{box}, x_{cls})$
\end{algorithm}

\vspace{0.3em}
\noindent
\textbf{Total time steps.} In the simulations conducted, the number of time steps $T$ utilized within the proposed ConsistencyDet framework is configured to 40. This figure represents a notable reduction from the 1000 time steps employed by DiffusionDet, illustrating a marked contrast in the temporal dynamics inherent in the inference processes of the two methodologies. The approach adopted in ConsistencyDet greatly reduces computation time and memory consumption, and may result in only a small decrease in accuracy. During the experiment, it was found that  accuracy did not decrease significantly, but increased in most experiments compared with DiffusionDet.

\vspace{0.3em}
\noindent
\textbf{Box-renewal.} Subsequent to each sampling iteration, the predicted outcomes are segregated into two categories: desired and undesired bounding boxes. Desired bounding boxes are those that are precisely aligned with the corresponding objects, whereas undesired bounding boxes remain in positions akin to a random distribution. Propagating these undesired bounding boxes to subsequent iterations is typically counterproductive, as their distributions may be quite different with the situations of padded GT boxes with additional noise during the training phase. Therefore, a Box-renewal operation is implemented both in the training and inference phases. This operation entails the discarding of undesired bounding boxes, identified by scores falling beneath a certain threshold. Their subsequent replacement with freshly sampled random boxes, is derived from a Gaussian distribution.

\section{Experiment}
\label{Expe}
In this section, we evaluate the performance of our model using two prevalent datasets: MS-COCO and LVIS, as cited in \cite{lin2014microsoft,gupta2019lvis}. Initially, we conduct experiments to ascertain the optimal values for key parameters in the MS-COCO and LVIS datasets.  Subsequently, we compare the proposed ConsistencyDet framework against a range of established detection models, including the Diffusion Model. Finally, we undertake ablation studies to dissect and analyze the individual components of the ConsistencyDet framework.

\vspace{0.3em}
\noindent
\textbf{MS-COCO} \cite{lin2014microsoft} comprises 118,000 training images in the train2017 subset and 5,000 validation images in the val2017 subset, spanning 80 different object categories. A performance evaluation is conducted using the standard metrics of box average precision (AP), and AP at IoU thresholds of 0.5 (AP\(_{50}\)) and 0.75 (AP\(_{75}\)). The IoU metric quantifies the proportionate overlap between each predicted bounding box and the corresponding GT boxes, offering important insight into the precision of the object detector at various levels of accuracy. Besides, scale-specific AP scores are also adopted: AP$_{s}$ / AP$_{m}$ / AP$_{l}$ for small / median / large objects separately.

\vspace{0.3em}
\noindent
\textbf{LVIS v1.0} \cite{gupta2019lvis} encompasses 100,000 training images and 20,000 validation images, drawing from the same source images as MS-COCO. It focuses on large-vocabulary object detection and instance segmentation by annotating a long-tailed distribution of objects across 1,203 categories. The evaluation framework for this dataset incorporates a suite of metrics: AP / AP\(_{50}\) / AP\(_{75}\), and AP$_{s}$ / AP$_{m}$ / AP$_{l}$ for small / median / large objects separately (scale-specific AP scores), AP$_{r}$ / AP$_{c}$ / AP$_{f}$ for rare / common / frequent categories separately (category-specific AP scores). These metrics provide a multifaceted evaluation of model performance across varying levels of detection precision and category frequency.

\subsection{Implementation detail}
\label{Implementation}
\noindent
During training, we initialize our model using pre-trained weights from ImageNet-1K for the ResNet backbone and ImageNet-21K \cite{deng2009imagenet} for the Swin-base backbone. The detection decoder in the ConsistencyDet architecture starts with Xavier initialization \cite{glorot2010understanding}. We utilize the AdamW optimizer \cite{loshchilov2017decoupled}, with an initial learning rate of $2.5 \times 10^{-5}$ and a weight decay of $10^{-4}$. Training employs mini-batches of size 8 across 4 GPUs. For MS-COCO, we follow a standard training schedule of 350,000 iterations, reducing the learning rate by a factor of ten at 90,000 and 280,000 iterations. For LVIS, reductions occur at 100,000, 300,000, and 320,000 iterations. We apply augmentation techniques, including random horizontal flipping, scale jitter resizing (with the shortest side between 480 and 800 pixels and the longest side not exceeding 1333 pixels), and random cropping. However, we do not utilize more robust augmentations like EMA, MixUp \cite{zhang2017mixup}, or Mosaic \cite{ge2021yolox}.

\begin{table}[]
\caption{Performance comparison of different object detectors on the COCO 2017 val set.}
\scalebox{1}{
    \begin{threeparttable}
        \begin{center}
        \begin{tabular}{@{}lllllll@{}}
            \toprule
            Method                  & AP            & AP$_{50}$     & AP$_{75}$     & AP$_s$       & AP$_m$        & AP$_l$        \\ \midrule
            \multicolumn{7}{c}{Res-Net50} \\

            Faster R-CNN~\cite{ren2016faster}             & 40.2          & 61.0            & 43.8          & 24.2          & 43.5          & 52.0            \\
            Cascade R-CNN~\cite{cai2019cascade}            & 44.3          & 62.2          & 48.0            & 26.6          & 47.7          & 57.7          \\
            DETR~\cite{carion2020end}                    & 42.0            & 62.4          & 44.2          & 20.5          & 45.8          & \underline{61.1}          \\

            Sparse R-CNN~\cite{sun2021sparse}             & 45.0            & 63.4          & 48.2          & 26.9          & 47.2          & 59.5          \\
            DiffusionDet~\cite{chen2023diffusiondet}    & \underline{46.2}          & \textbf{66.4}          & \underline{49.5}          & \underline{28.7}          & \underline{48.5}          & \textbf{61.5}          \\
            \textbf{ConsistencyDet} & \textbf{46.9} & \underline{65.7} & \textbf{51.3} & \textbf{30.2} & \textbf{50.2} & 61.0 \\ \midrule
            \multicolumn{7}{c}{Res-Net101} \\

            Faster R-CNN~\cite{ren2016faster}             & 42.0            & 62.5          & 45.9          & 25.2          & 45.6          & 54.6          \\
            Cascade R-CNN~\cite{cai2019cascade}            & 45.5          & 63.7          & 49.9          & 27.6          & 49.2          & 59.1          \\
            DETR~\cite{carion2020end}                     & 43.5          & 63.8          & 46.4          & 21.9          & 48.0            & 61.8          \\
            Sparse R-CNN~\cite{sun2021sparse}             & 46.4          & 64.6          & 49.5          & 28.3          & 48.3          & 61.6          \\
            DiffusionDet~\cite{chen2023diffusiondet}     & \underline{47.1}          & \textbf{67.1}          & \underline{50.6}          & \underline{30.2}          & \underline{49.8}          & \textbf{62.7}         \\
            \textbf{ConsistencyDet} & \textbf{47.2} & \underline{66.8} & \textbf{50.9} & \textbf{31.3} & \textbf{50.4} & \underline{62.3} \\   \midrule
            \multicolumn{7}{c}{Swin-base} \\
            Cascade R-CNN~\cite{cai2019cascade}            & 51.9          & 70.9          & 56.5          & 35.4          & 55.2          & 51.9          \\
            Sparse R-CNN~\cite{sun2021sparse}              & 52.0            & 72.2          & \underline{57.0}            & 35.8          & 55.1          & 52.0            \\
            DiffusionDet~\cite{chen2023diffusiondet}     & \underline{52.8}          & \textbf{73.6}          & 56.8          & \underline{36.1}          & \underline{56.2}          & \textbf{68.8}          \\
            \textbf{ConsistencyDet} & \textbf{53.0} & \underline{73.2} & \textbf{57.6} & \textbf{36.5} & \textbf{57.2} & \underline{68.4} \\
            \bottomrule
        \end{tabular}
			\begin{tablenotes}
				\footnotesize
				\item[1]Results of the above evaluation metrics are all percentage data (\%).
                \item[2]Bold font indicates the best performance while  underlined font indicates the second best.
			\end{tablenotes}

        \end{center}
        \end{threeparttable}

}
\label{tab:3}
\end{table}

\begin{table}[]
\caption{Performance comparison of different object detectors on LVIS v1.0 val set.}
    {
    \scalebox{1}{
        \begin{threeparttable}
        \begin{tabular}{@{}lcccccc@{}}
            \toprule
                Method                  & AP            & AP$_{50}$     & AP$_{75}$     & AP$_s$        & AP$_m$      & AP$_l$            \\ \midrule
                \multicolumn{7}{c}{Res-Net50} \\
                Faster R-CNN~\cite{ren2016faster}             & 25.2          & 40.6          & 26.9          & 18.5          & 32.2        & 37.7                 \\
                Cascade R-CNN~\cite{cai2019cascade}            & 29.4          & 41.4          & 30.9          & 20.6          & 37.5        & 44.3           \\
                Sparse R-CNN~\cite{sun2021sparse}             & 29.2          & 41.0            & 30.7          & 20.7          & 36.9        & 44.2                \\
                DiffusionDet~\cite{chen2023diffusiondet}     & \underline{31.9}          & \textbf{45.3}          & \underline{33.1}          & \underline{22.8}          & \underline{40.2 }       & \underline{48.1}       \\

                \textbf{ConsistencyDet} & \textbf{32.2} & \underline{45.1} & \textbf{33.7} & \textbf{23.1} & \textbf{40.4} & \underline{47.7} \\ \midrule
                \multicolumn{7}{c}{Res-Net101} \\
                Faster R-CNN~\cite{ren2016faster}             & 27.2          & 42.9          & 29.1        & 20.3          & 35            & 40.4            \\
                Cascade R-CNN~\cite{cai2019cascade}            & 31.6          & 43.8          & 33.4        & 22.3          & 39.7          & 47.3             \\
                Sparse R-CNN~\cite{sun2021sparse}             & 30.1          & 42.0            & 31.9        & 21.3          & 38.5          & 45.6             \\
                DiffusionDet~\cite{chen2023diffusiondet}    & \textbf{33.5}          & \textbf{47.3}          & \underline{34.7}        & \textbf{23.6}          & \textbf{41.9}          & \textbf{49.8}     \\

                \textbf{ConsistencyDet} & \underline{33.1} & \underline{46.}1 & \textbf{34.8} & \underline{22.9} & \underline{41.5} & \underline{49.4} \\ \midrule
                \multicolumn{7}{c}{Swin-base} \\

                DiffusionDet~\cite{chen2023diffusiondet}     & \underline{42.1}          & \textbf{57.8}          & \underline{44.3}          & \textbf{31.0}            & \underline{51.3}          & \underline{62.5}           \\

                \textbf{ConsistencyDet} & \textbf{42.4} & \underline{57.1} & \textbf{44.9} & \underline{30.3} & \textbf{51.5} & \textbf{63.4} \\

            \bottomrule
        \end{tabular}
			\begin{tablenotes}
				\footnotesize
				\item[1]Results of the above evaluation metrics are all percentage data (\%).
                \item[2]Bold font indicates the best performance while underlined font indicates the second best performance.
			\end{tablenotes}
        \end{threeparttable}
        }
        }

\label{tab:4}
\end{table}


\subsection{Main Property}
\label{Quantitative}

\begin{figure*}[!t]
\centering
\includegraphics[width=0.95\linewidth]{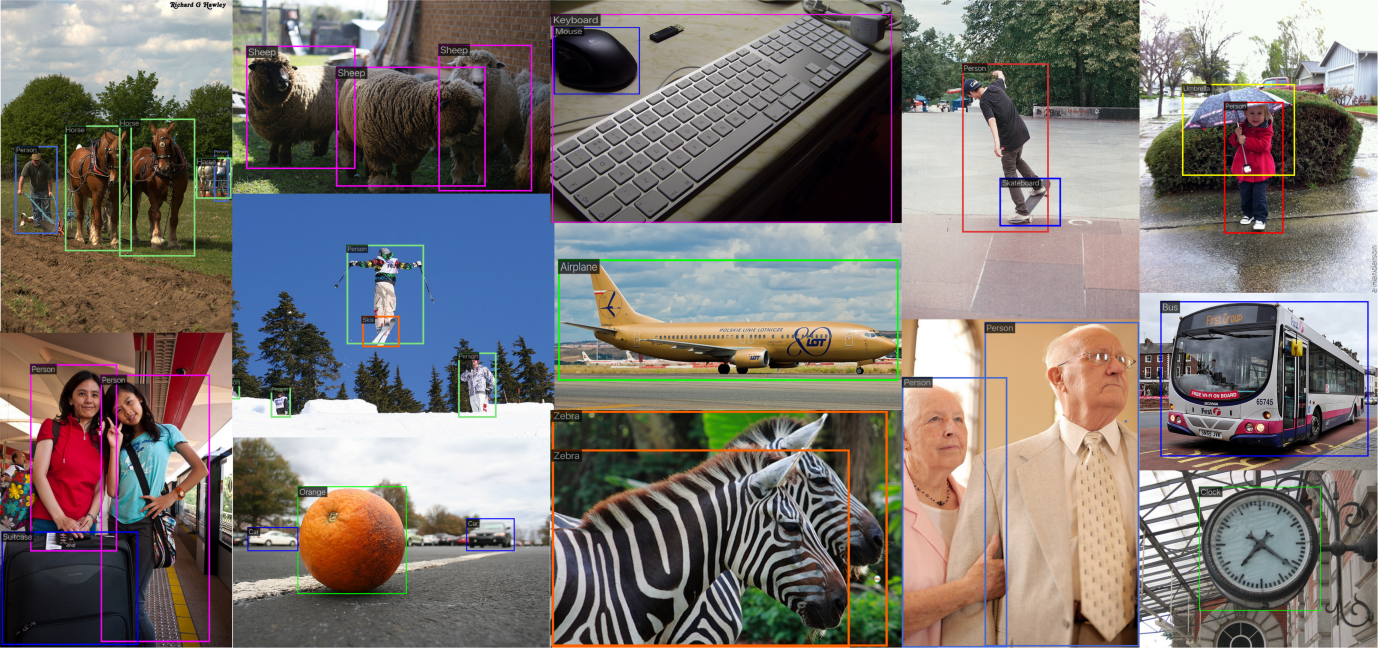} 
\centering
\caption{Typical instances of detected objects by the proposed ConsistencyDet. The objects in these images are detected with correct categories and accurate borders.}
\label{fig:9}
\end{figure*}

The key characteristic of ConsistencyDet is self-consistency, which stabilizes the mapping from any point along the time axis back to the origin. This stability provides flexibility during inference, allowing for adjustments in the number of sampling steps. Increasing the sampling steps can improve detection accuracy, although it may reduce computational efficiency. Consequently, the model can balance accuracy and efficiency based on specific application needs. Additionally, ConsistencyDet exhibits strong noise resistance, allowing for dynamic adjustments in the number of bounding boxes during inference. Some detection results are shown in Fig. \ref{fig:9}.

\begin{figure}[!t]
\centering
\includegraphics[width=0.95\linewidth]{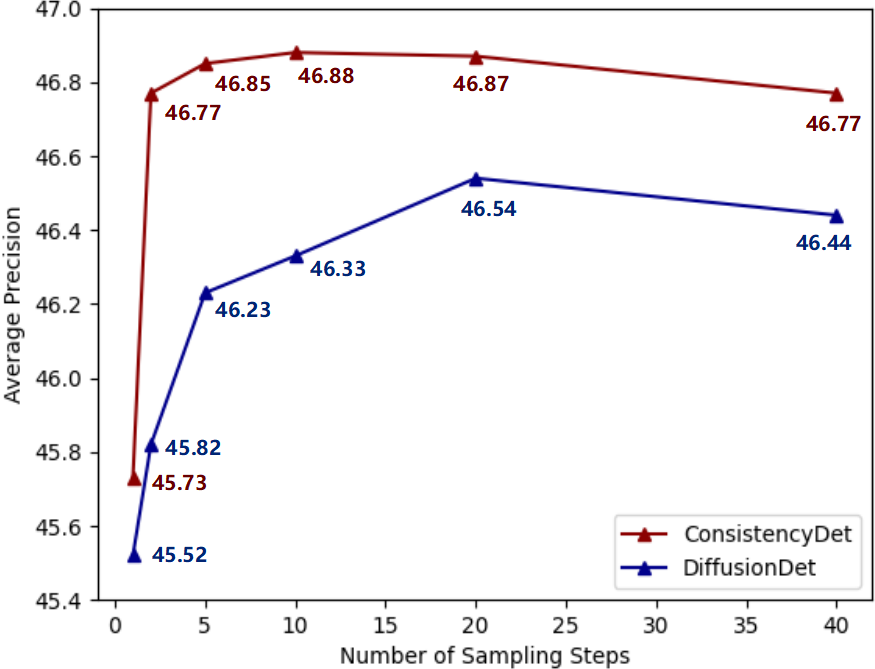}
\caption{Performance analysis of progressive refinement. ConsistencyDet is evaluated with varied sampling time steps. It is trained on the MS-COCO dataset with ResNet-50 as the backbone, using 500 proposal boxes for evaluation at each sampling time step. Simultaneously, the accuracy of ConsistencyDet surpasses that of DiffusionDet under the same experimental parameters. Here, the results of AP are all percentage data (\%).}
\label{fig:3}
\end{figure}

\vspace{0.3em}
\noindent
\textbf{Refinement Through Iterative Sampling.}
For inference, the number of proposal boxes is fixed at 500 at each sampling step. In the MS-COCO dataset, ConsistencyDet shows stable performance improvements as sampling steps increase from 1 to 10, as illustrated in Fig. \ref{fig:3}. The AP for ConsistencyDet rises from 45.73\% at $T=1$ to 46.88\% at $T=10$, indicating that higher accuracy can be achieved with optimal sampling steps.

\begin{figure}[!t]
\centering
\includegraphics[width=0.95\linewidth]{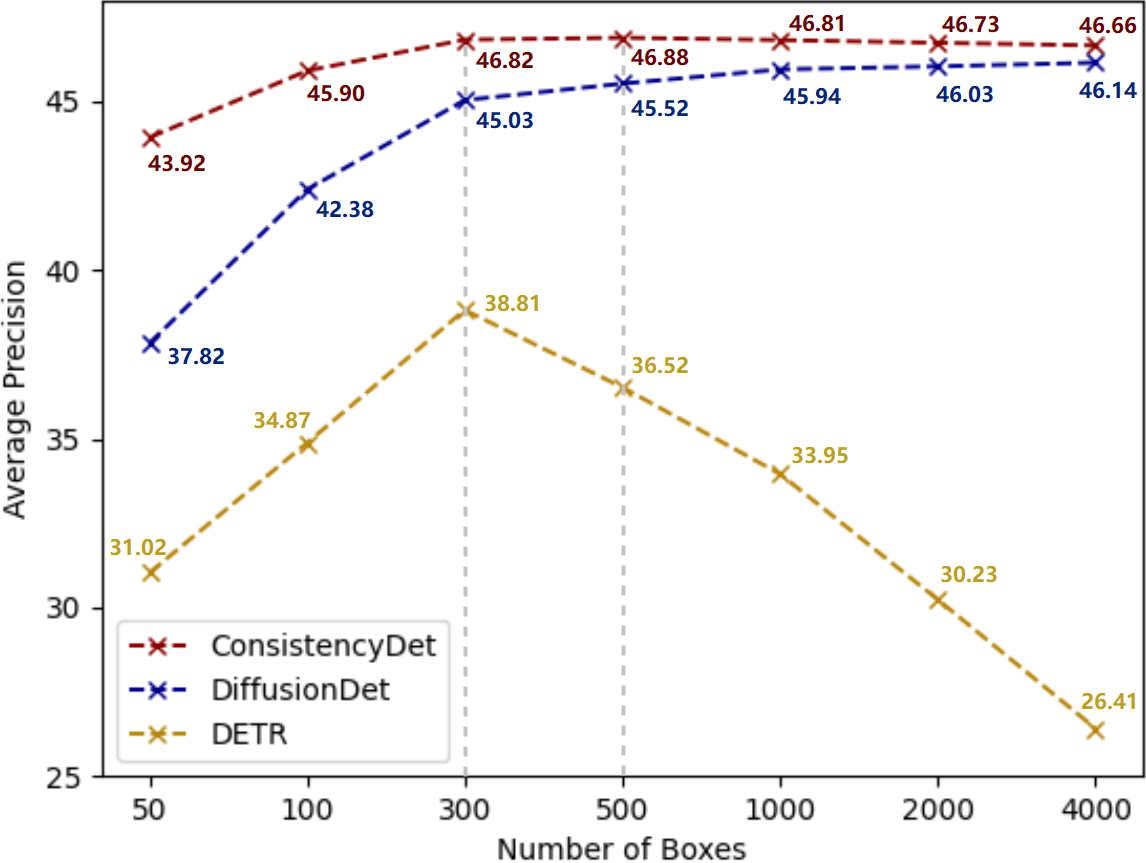}
\caption{Performance comparison of dynamic box sampling for improved robustness. More proposal boxes in inference lead to accuracy improvement at first and tend towards stability for both ConsistencyDet and DiffusionDet, while causing a degradation in DETR due to its lack of noise resistance. Furthermore, ConsistencyDet outperforms DiffusionDet for all these cases. Here, the results of AP are all percentage data (\%).}
\label{fig:ts}
\end{figure}

\vspace{0.3em}
\noindent
\textbf{Dynamic Box Sampling for Improved Robustness.}
The efficacy of ConsistencyDet is evaluated against DiffusionDet and DETR on the MS-COCO dataset, as shown in Fig. \ref{fig:ts}. ConsistencyDet improves performance significantly as the number of boxes increases, stabilizing when $n_p > 300$ and achieving optimal performance at $n_p = 500$. In contrast, DETR peaks at $n_p = 300$, then declines sharply, dropping to 26.4\% at $n_p = 4000$, a 12.4\% reduction from its peak. At the same time, ConsistencyDet outperforms DiffusionDet for all cases. Moreover, this experimental results demonstrate that ConsistencyDet significantly surpasses traditional detection methods like DETR in terms of noise resistance.

\begin{figure*}[!t]
\centering
\includegraphics[height=3.77in]{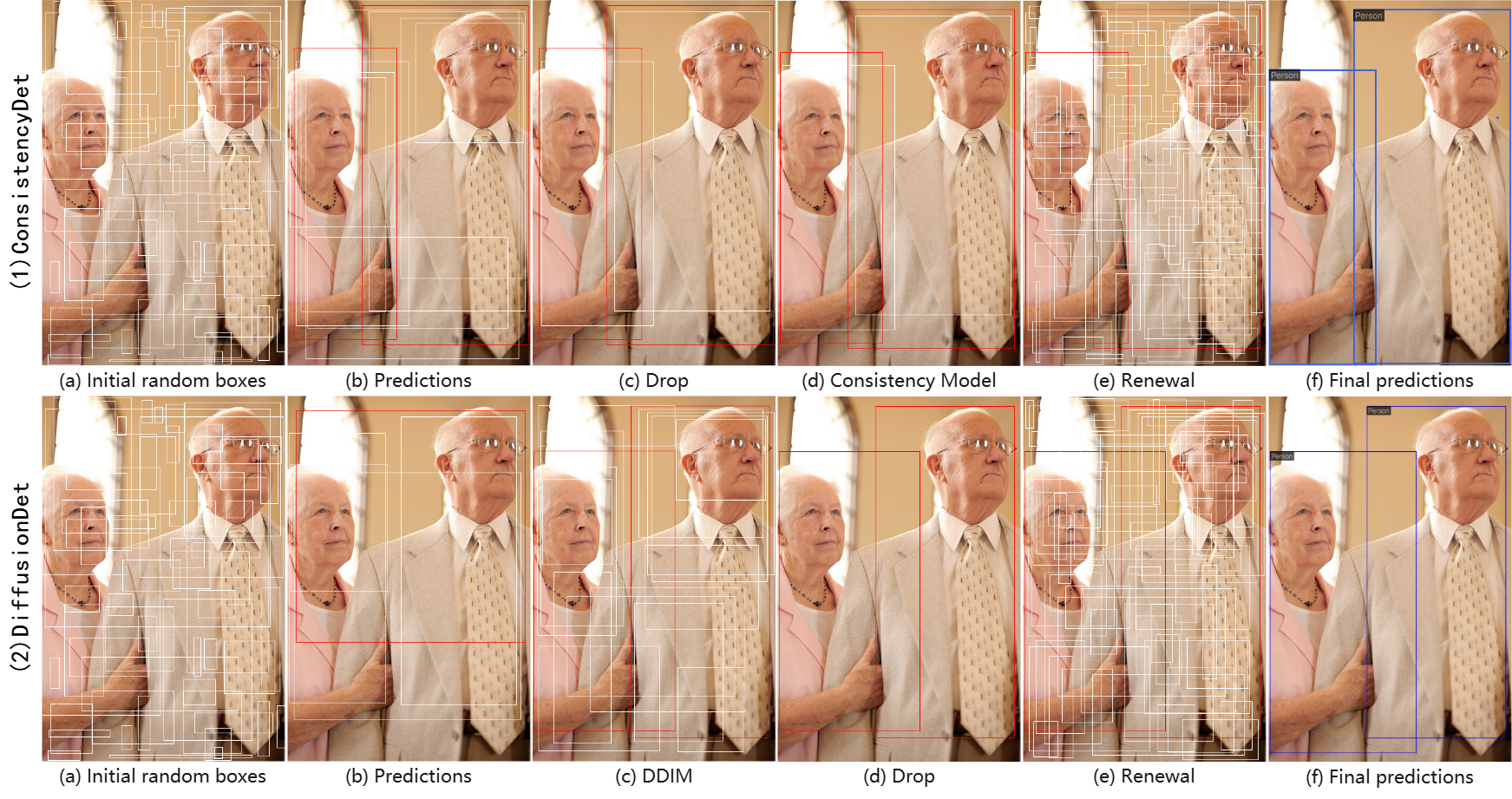} 
\centering
\caption{A comparison of the visual reasoning process with one typical sampling step between ConsistencyDet and DiffusionDet. We selected a sample image from the COCO 2017 val set for testing with both ConsistencyDet and DiffusionDet. The initial noised boxes or verified boxes with low confidence are marked in white, while the boxes with high confidence are marked in red and the final predictions are marked in blue.}
\label{fig:ts2}
\end{figure*}

\vspace{0.3em}
\noindent
\textbf{Sampling step.} 
As depicted in Fig. \ref{fig:ts2}, the sampling phase starts with the initialization of noised boxes at the $t$-th time step, with the noise addition protocol. These noised boxes are then put into the model to generate predictions. The detection decoder predicts the category scores and box coordinates of the current step. An initial filtration process ensues, whereby only the bounding boxes surpassing a predetermined confidence threshold are preserved. Under the theoretical guidance of the Consistency Model, noised boxes are predicted, optimized and filtered for the next time step. At the end of each time step, additional proposals are supplemented by the Box-renewal operation, ensuring the total amount of noise boxes reaches $n_p=500$ for the next iteration. In the final time step, NMS is applied to refine the results. By comparing the detection results in Fig. \ref{fig:ts2}(f), it is evident that ConsistencyDet performs better than DiffusionDet, with more precise borders.

\subsection{Simulation analysis}
\label{Qualitative}
The performance of ConsistencyDet is evaluated against other state-of-the-art detector models \cite{ren2016faster,cai2019cascade,sun2021sparse,carion2020end,chen2023diffusiondet} on the MS-COCO and LVIS v1.0 datasets. The following provides an analysis of the simulation results.

\vspace{0.3em}\noindent
\textbf{MS-COCO.} Performance metrics for different detectors are summarized in~\cref{tab:3}. ConsistencyDet, with a ResNet-50 backbone, achieves an AP of 46.9\%, outperforming Faster R-CNN, DETR, Sparse R-CNN, and DiffusionDet. Upgrading to a ResNet-101 backbone boosts ConsistencyDet's AP to 47.2\%, exceeding these baselines. Additionally, with a Swin-base backbone \cite{liu2021swin} pretrained on ImageNet-21k \cite{deng2009imagenet}, ConsistencyDet achieves an impressive AP of 53.0\%, outperforming all compared detectors. This shows our method's consistent performance improvement with varied backbones, confirming its effectiveness in detection accuracy.

\vspace{0.3em}
\noindent
\textbf{LVIS v1.0.} ~\cref{tab:4} compares the object detection performance of ConsistencyDet with several other models. Using a ResNet-50 backbone, reproductions of Faster R-CNN and Cascade R-CNN yielded APs of 25.2\% and 29.4\%, respectively, while ConsistencyDet achieved 32.2\%, outperforming Faster R-CNN, DETR, and Sparse R-CNN and remaining highly competitive with DiffusionDet. With the ResNet-101 backbone, ConsistencyDet reaches an AP of 33.1\%, slightly below DiffusionDet but above other baselines. When using the Swin-base backbone \cite{liu2021swin}, ConsistencyDet achieves 42.4\% AP, outperforming all other mentioned baselines.

\subsection{{Ablation Study}}
\label{discussion}

Comprehensive ablation studies were carefully conducted to further elucidate the key characteristics of the proposed ConsistencyDet model on the MS-COCO and LVIS v1.0 datasets. These simulations utilized a ResNet-50 architecture equipped with a FPN as the primary backbone, with no additional modifications or enhancements specified.

\vspace{0.3em}\noindent
\textbf{Box Renewal threshold and NMS threshold.}
The left column of~\cref{tab:5} illustrates the impact of the score threshold, denoted as $B_{th}$, on the Box-renewal process. A threshold value of 0.0 implies that Box-renewal is not employed. According to the evaluation on the COCO 2017 validation set, employing a threshold of 0.98 marginally outperforms other threshold settings.The right column of~\cref{tab:5} delineates the effects of varying the NMS score threshold, represented as $N_{th}$, on the AP. Setting the threshold to 0.0 signifies the absence of any NMS procedure. An analysis of the COCO 2017 validation set suggests that a threshold of 0.64 yields a performance that is modestly superior relative to the other thresholds investigated.

\begin{table}[]

\caption{Optimal choices for the thresholds for the Box-renewal and NMS operation.}
\centering
\scalebox{0.9}{    \begin{threeparttable}
        \begin{tabular}{cccc|cccc}
            \toprule
            $B_{th}$ & AP & AP$_{50}$ & AP$_{75}$ & $N_{th}$ & AP & AP$_{50}$ & AP$_{75}$\\ \midrule
            - & - & - & - & 0.5 & 46.5 & \textbf{66.6} & 50.0\\
            0.9 & 46.5 & 65.5 & 50.9 & 0.6 & 46.8 & 66.0 & 50.8\\
            0.98 & \textbf{46.9} & \textbf{65.7} & \textbf{51.3} & 0.64 & \textbf{46.9} & 65.7 & 51.3\\
            1.0 & 46.7 & 65.6 & 51.2 & 0.7 & 46.6 & 64.8 & \textbf{51.5}\\
            \bottomrule
        \end{tabular}
			\begin{tablenotes}
				\footnotesize
				\item[1]Results of AP \!/\! AP$_{50}$ \!/\! AP$_{75}$ are all percentage data (\%).
                \item[2]Bold fonts indicate the best performance.
			\end{tablenotes}
        \end{threeparttable}}

\label{tab:5}
\end{table}

\begin{figure}[!t]
\centering
\includegraphics[width=3.5in]{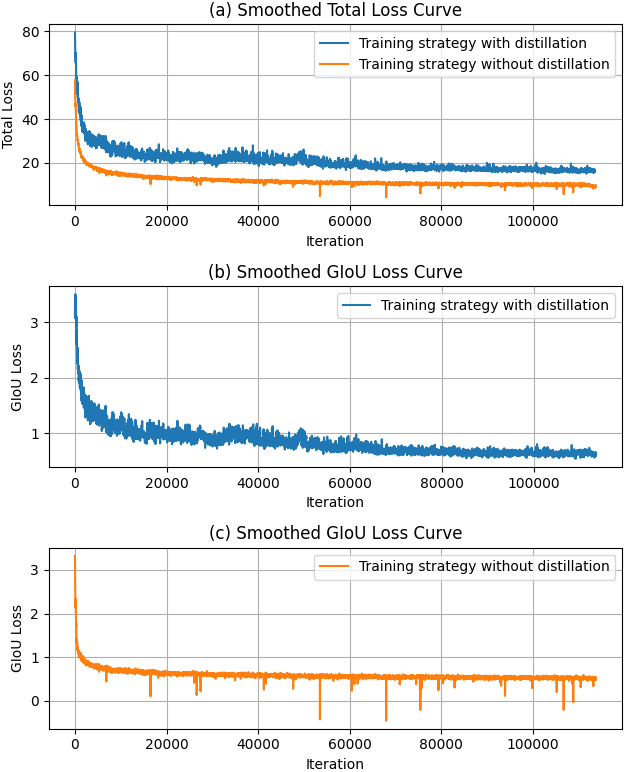} 
\centering
\caption{Smoothed loss curves of ConsistencyDet via different training strategies. (a) The comparison of the total loss curves between two training strategies. (b) Smoothed GIoU loss curve of training strategy with distillation. (c) Smoothed GIoU loss curve of independent training strategy without distillation.}
\label{fig:loss}
\end{figure}

\vspace{0.3em}
\noindent
\textbf{Training strategy.} Algorithm \ref{alg:4} provides two training strategies, differentiated by whether to utilize a pre-trained DiffusionDet model for the distillation training of ConsistencyDet. In Fig. \ref{fig:loss}, the loss curves of these two strategies are compared in the coordinate system. Both are trained with 4 GPUs, and the batch size is set to 2 for each GPU. The training process adopts Res-Net50 as the backbone and is conducted on the MS-COCO dataset. While DiffusionDet could, in theory, achieve better results with an adequate number of denoising time steps, the complexity of its operational steps presents challenges in effective training, which in turn limits its performance in practice. Hence, our experimental efforts primarily focus on the training strategy of the Consistency Model in isolation. From Fig. \ref{fig:loss}(b)-(c), it is evident that employing distillation training can effectively mitigate the issue of outliers encountered during the training process. At the latter half of the timeline, the stochastic nature introduced by noise may sometimes lead to unreasonable values of box coordinates, where the top-left coordinates are smaller than the bottom-right coordinates. This illogical issue results in negative GIoU loss values, potentially causing the model to converge in an erroneous direction, which could pose potential risks. Therefore, the introduction of distillation brings corresponding benefits. Since the predictions of current DiffusionDet still have a big gap with GT information, if the position information of the boxes predicted by DiffusionDet is used as the reference information from the origin of the timeline, this operation inevitably leads to inadequate guidance in the training process of ConsistencyDet, and results in a larger total loss with poor performance, shown as Fig. \ref{fig:loss}(a). Therefore, after comprehensive consideration, this work adopts the independent training strategy for the training process which achieves better results. Also, the aforementioned illogical issue can be restrained with the conservative limitation of value ranges.

\vspace{0.3em}
\noindent
\textbf{Accuracy vs Speed.}
The inference speeds of ConsistencyDet and DiffusionDet, both utilizing a ResNet-50 backbone, are compared in~\cref{tab:9}. Runtime performance was measured on a single RTX3080 GPU with a batch size of one. DiffusionDet's evaluation includes sampling timesteps ($t_{ss} \!\!=\!\! 2/10/20/40$), recording both AP and FPS. ConsistencyDet was tested at total timesteps \(t_{ss} = 2/10/20\). The experimental results show that ConsistencyDet reaches its optimal detection accuracy at \(t_{ss} = 10\), while at \(t_{ss} = 2\), its accuracy is only 0.1\% lower than the peak performance, aligning well with the ``few-step denoising'' mechanism of ConsistencyDet. In contrast, DiffusionDet requires \(t_{ss} = 20\) to achieve its best detection accuracy, but even at this setting, it performs worse than ConsistencyDet at \(t_{ss} = 2\) and is 7.2 times less efficient in inference speed. This difference clearly demonstrates the significant advantages of ConsistencyDet in both speed and accuracy. Moreover, ConsistencyDet consistently maintains superior detection accuracy at all sampling steps compared to DiffusionDet, further underscoring its stability. In summary, our method achieves stable and enhanced detection performance with only a few sampling steps, overcoming the main drawback of DiffusionDet, which sacrifices inference speed to improve detection accuracy.


\begin{table}[]
\centering
\caption{Accuracy vs. speed comparison of ConsistencyDet and DiffusionDet at different sampling time steps.}
\setlength{\tabcolsep}{10pt}
\scalebox{0.95}{
\begin{threeparttable}
    \begin{tabular}{@{}lccc@{}}
        \toprule
        Model & $n_{ss}$ & AP & FPS\\
        \midrule
        \multirow{4}{*}{\centering DiffusionDet}
        & 2 & 45.82 & 6.734\\
        & 10 & 46.33 & 1.987\\
        & 20 & 46.54 & 0.954\\
         & 40 & 46.44 & 0.473\\ \midrule
        \multirow{2}{*}{\centering ConsistencyDet}  & 2 & 46.8 & \textbf{6.892} \\
         & 10 & \textbf{46.88} & 1.783 \\ & 20 & 46.87 & 0.845 \\ \bottomrule
    \end{tabular}
    			\begin{tablenotes}
				\footnotesize
				\item[1]Results of AP are all percentage data (\%).
                \item[2]Bold font indicates the best performance.
			\end{tablenotes}
    \end{threeparttable}
}

\label{tab:9}
\end{table}

\section{Conclusion}
\label{Conc}

In this work, we have introduced a novel and efficient object detection paradigm, ConsistencyDet, which conceptualizes object detection as a denoising diffusion process evolving from noisy boxes to precise targets. The proposed ConsistencyDet leverages the Consistency Model for object detection and presents an innovative and robust strategy for the addition of noise and denoising. A key feature of the ConsistencyDet is its self-consistency, which ensures that noise information at any temporal stage can be directly mapped back to the origin on the coordinate axis, thus significantly boosting the algorithm's computational efficiency and overall performance.

Through extensive simulations conducted on standard object detection benchmarks, it has been confirmed that ConsistencyDet achieves competitive performance in comparison to other established detectors. In a head-to-head comparison with DiffusionDet, ConsistencyDet displays efficient processing across these datasets. Notably, with the selection of optimal sampling time steps for both models, ConsistencyDet demonstrates substantial gains in computational efficiency over DiffusionDet. This result is a testament to the self-consistency feature intrinsic to ConsistencyDet.

Prospective enhancements to ConsistencyDet are manifold and could serve to further refine the approach. Future work may include: (1) Revising noise addition and denoising techniques to improve detection accuracy, particularly for large objects. (2) Expanding the performance capabilities of the Consistency Model in object detection tasks, potentially through the implementation of training strategies that incorporate knowledge distillation from well-trained Diffusion Models. (3) Enhancing computational efficiency by simplifying the operational steps in the model. (4) Extending the proposed noise addition and denoising framework to additional domains, such as image segmentation and object tracking.

\bibliographystyle{IEEEtran}
\bibliography{ConsistencyDet}

\end{document}